\documentclass[3p,times,procedia]{elsarticle}

\usepackage{ecrc}

\volume{00}

\firstpage{1}

\journalname{Procedia Computer Science}

\runauth{}

\jid{procs}

\jnltitlelogo{Procedia Computer Science}

\CopyrightLine{2011}{Published by Elsevier Ltd.}

\usepackage{amssymb}

\usepackage[figuresright]{rotating}
\usepackage{comment}
\usepackage[hang,tight]{subfigure}
\usepackage{amsmath}
\usepackage{bm}
\usepackage{booktabs}
\usepackage{multirow}

\usepackage{algorithm}
\usepackage{algorithmic}
\usepackage{color}
\usepackage{caption}

\begin{document}

\begin{frontmatter}



\dochead{}


\title{Boundary-Enhanced Time Series Data Imputation with Long-Term Dependency Diffusion Models}


\author[label1]{Chunjing Xiao }
\ead{xiaocj@henu.edu.cn}

\author[label1]{Xue Jiang}
\ead{jiangxue@henu.edu.cn}

\author[label1]{Xianghe Du}
\ead{xianghedu@henu.edu.cn}

\author[label1]{Wei Yang}
\ead{yangwei@henu.edu.cn}

\author[label2]{Wei Lu\corref{cor1}}
\cortext[cor1]{Corresponding author.}
\ead{luwei@wmu.edu.cn}

\author[label3]{Xiaomin Wang}
\ead{xmwang@uestc.edu.cn}

\author[label4]{Kevin Chetty}
\ead{k.chetty@ucl.ac.uk}


\address[label1]{The School of Computer and Information Engineering, Henan University, Kaifeng, 475004 China}
\address[label2]{Quzhou Affiliated Hospital of Wenzhou Medical University, Quzhou People's Hospital, Quzhou 324000, China.}
\address[label3]{Yangtze Delta Region Institute (Quzhou), University of Electronic Science and Technology of China, Quzhou 324000, China.}
\address[label4]{The Department of  Security and Crime Science,  University College London, London WC1E 6BT, UK}


\begin{abstract}  
Data imputation is crucial for addressing challenges posed by missing values in multivariate time series data across various fields, such as healthcare, traffic, and economics, and has garnered significant attention. Among various methods, diffusion model-based approaches show notable performance improvements. However, existing methods often cause disharmonious boundaries between missing and known regions and overlook long-range dependencies in missing data estimation, leading to suboptimal results. To address these issues, we propose a Diffusion-based time Series Data Imputation (DSDI) framework. We develop a weight-reducing injection strategy that incorporates the predicted values of missing points with reducing weights into the reverse diffusion process to mitigate boundary inconsistencies. Further, we introduce a multi-scale S4-based U-Net, which combines hierarchical information from different levels via multi-resolution integration to capture long-term dependencies. Experimental results demonstrate that our model outperforms existing imputation methods.
\end{abstract}



\end{frontmatter}



\newcommand{\modelname}{DSDI}

\section{Introduction}
\label{sec:introduction}
Time series data is prevalent across various fields, such as traffic, economics, healthcare, and meteorology. However, in multivariate time series data, the problem of missing values arises when certain time points or sensor measurements are absent from the data. This issue is prevalent in real-world scenarios and can result from sensor malfunctions, data collection inaccuracies, network communication problems, and other factors~\cite{wu2022data,adhikari2022comprehensive}. Missing values can persist for extended periods under challenging environments (e.g., polar regions) where personnel may not be able to address hardware-related issues with sensors promptly. 
%
The presence of missing values hinders the ability to achieve high accuracy and reliability in data analysis and forecasting~\cite{zheng2024graph,miao2022experimental}.
For example, in medical care, the accuracy of diagnosing a patient’s cancer relies heavily on the specialist’s experience. However, missing patient data can significantly impact the decisions made by specialists~\cite{wu2018decision}.
To effectively learn from such incomplete data, handling the missing portions before modeling is an unavoidable step.


Data imputation is an effective method for handling missing data, enabling the estimation or completion of missing values using known data and facilitating the analysis of complete datasets~\cite{wang2024time}. This minimizes the influence of missing data on subsequent analyses, resulting in more precise research outcomes. Owing to the growing need for accurate data analyses, data imputation has garnered significant interest~\cite{ma2020end}. Initially, statistical methods and matrix factorization are introduced for handling missing data~\cite{amiri2016missing,chen2019bayesian}. However, these methods fall short in capturing the intricate temporal relationships and complex variation patterns inherent in time series data, resulting in limited performance~\cite{wang2024deep}. 
Recently, deep learning techniques have been explored for time series data imputation, including approaches based on Recurrent Neural Networks (RNNs), Variational AutoEncoders (VAEs), and Generative Adversarial Networks (GANs)~\cite{adhikari2022comprehensive}.


Due to the unstable training process of these methods such as VAEs and GANs~{\cite{du2023saits, wu2020attention}, diffusion models have recently emerged as an alternative for imputing time series data~\cite{alcaraz2023diffusionbased, tashiro2021csdi, lin2023diffusion}. Diffusion models can generate diverse plausible imputations by converting noise into a clean data sample through iterative denoising, outperforming counterparts such as RNNs, VAEs, and GANs~\cite{wang2023observed}. Despite this remarkable progress, existing diffusion model-based imputation methods still have some drawbacks: (1) They may distort the boundaries between missing and known regions, particularly during the initial phase of the reverse process, leading to inferior performance. Since the reverse diffusion starts with noise, the generated values for missing points in the initial phase contain significant noise. This discrepancy creates disharmonious boundaries because known points retain realistic values, resulting in a mismatch with the noisy values of missing points. Such disharmony can significantly deteriorate imputation performance~\cite{lugmayr2022repaint, shen2023non}. (2) Typical diffusion model-based methods might suffer from information loss during downsampling and the limited receptive fields of convolutional layers, which hinder the model's ability to capture long-range dependencies for missing data estimation. Although U-Net is frequently employed as the denoising network in general diffusion models~\cite{ronneberger2015u}, the convolutional and pooling operations in U-Net may not effectively capture long-range dependencies~\cite{gu2022efficiently, gu2022parameterization}. Therefore, in scenarios with continuous missing points, it is essential to exploit long-range dependencies to capture information beyond the missing region for accurate data imputation.

To address these issues, we propose a Diffusion model-based time Series Data Imputation (\modelname) framework. In this framework, we estimate the values of missing points based on known values using the diffusion model~\cite{ho2020denoising}. During this process, we introduce a weight-reducing injection strategy to adaptively inject the predicted values of missing points into the reverse diffusion process, mitigating the problem of disharmonious boundaries. Moreover, we present a multi-scale temporal structured state-space sequence (S4)-based U-Net to integrate information from multiple levels at different resolutions, effectively capturing long-range dependencies.

In particular, the weight-reducing injection strategy first utilizes a linear autoregressive (AR) model to predict the values of missing points and then injects them into the values generated by the diffusion model. The injection weight gradually declines as the reverse diffusion iteration progresses. In the early stages, the predicted values from the AR model are often more accurate than the generated values, as the latter contain considerable noise. At this stage, the injection is assigned a larger weight to provide more accurate information for missing points, narrowing the gap between the values of missing points and known points, and alleviating their disharmony. As the reverse diffusion process continues, the generated values progressively approach realistic values, becoming more accurate than the predicted values. Hence, the injection in the later stages is assigned a lower weight. Consequently, this weight-reducing injection strategy can efficiently mitigate disharmony and enhance imputation performance.

To capture long-range dependencies, we devise a multi-scale TemS4-based U-Net to enhance imputation performance for large time-span data. While traditional diffusion models like DDPMs~\cite{ho2020denoising} utilize U-Net for denoising~\cite{ronneberger2015u}, its convolution and pooling operations are ineffective at capturing long-term dependencies~\cite{gu2022efficiently, gu2022parameterization}. Conversely, the S4 model~\cite{gu2022efficiently} has demonstrated success in capturing long-range dependencies across various domains, including audio waveforms, movie clips, and NLP~\cite{gu2022efficiently, gu2022parameterization, islam2022long}. To efficiently apply the S4 layer to time series data, we introduce the temporal S4 into U-Net for \modelname. This approach combines hierarchical information from different levels via multi-resolution integration and employs skip connections that link different layers to capture long-term dependencies and mitigate the impact of consecutive missing points. By integrating temporal S4 into the U-Net architecture, \modelname~effectively captures long-term dependencies and addresses the challenges posed by consecutive missing points.

The main contributions are as follows:
\begin{itemize}
    \item We design a diffusion model-based time series data imputation framework that introduces a weight-reducing injection strategy to adaptively incorporate predicted values of missing points into the reverse diffusion process, alleviating disharmonious boundaries.
    \item We propose a multi-scale TemS4-based U-Net to integrate hierarchical information through multi-resolution fusion, enabling the capture of long-range dependencies and improving imputation performance.
    \item Extensive experiments conducted on real-world datasets demonstrate that \modelname~outperforms several existing imputation methods.
\end{itemize}

\section{Preliminaries}
\label{sec:Preliminaries}


\subsection{Problem Definition}

Following commonly used notations, we use bold uppercase letters, bold lowercase letters, uppercase letters, and lowercase letters to denote matrices (e.g., $\mathbf{I}$), vectors (e.g., $\mathbf{x}$), scalars (e.g., $L$), and indices (e.g., $t$), respectively. In the context of imputing missing data, the problem can be formulated as follows: Given a multivariate time series dataset of length $L$, denoted as $\mathbf{x}^{0} = \{C^{0}_1, C^{0}_2, C^{0}_3, \dots, C^{0}_L\}$, where $C^{0}_l \in \mathbb{R}^D$ represents the observation at time $l$. Additionally, a zero-one indicator matrix $\mathbf{I}$ is constructed as a mask to denote the positions of missing values. Specifically, the elements of $\mathbf{I}$ are 0 when the corresponding value in $\mathbf{x}^0$ is missing and 1 otherwise. The objective of filling in missing values in multivariate time series data is to estimate the absent values.

\subsection{Diffusion Model}

A $T$-step denoising diffusion probabilistic model (DDPM) ~\cite{ho2020denoising} consists of two processes: the diffusion process with steps $t \in \{0,1,...,T\}$ and the reverse process $t \in \{T,T-1, ... ,0\}$. Given the data $\mathbf{x}^0$, the diffusion process in the first step $(t=0)$, $q(\mathbf{x}^0)$, is defined as the data distribution $\mathbf{x}^0$ on $\mathbb{R}^L$, where $L$ is the signal length in samples.

The DDPM method utilizes a diffusion process to transform the data ${\mathbf{x}^{0}}$ into white Gaussian noise ${\mathbf{x}^{T}} \sim {\mathcal{N}}(0,1)$ over $T$ time steps. Each step in the forward direction is determined by the following process:
\begin{equation}
    \begin{aligned}
        q(\mathbf{x}^{t}|\mathbf{x}^{t-1}) = \mathcal{N}(\mathbf{x}^{t};\sqrt{1-{\beta_t}}{\mathbf{x}^{t-1}},{\beta_t}\mathbf{I})
        \label{equ:diffusionProcess}.
    \end{aligned}
\end{equation}
At each time step $t$, the sample $\mathbf{x}^t$ is obtained by adding independently and identically distributed (i.i.d.) gaussian noise with variance $\beta_t$ to the previous sample $\mathbf{x}^{t-1}$. The previous sample is then scaled by $\sqrt{1-\beta_t}$ according to a predefined variance schedule.

The reverse process is represented by a neural network that estimates the parameters $\mu_\theta(\mathbf{x}^t,t)$ and $\sum_\theta(\mathbf{x}^t,t)$ of a Gaussian distribution: 
\begin{equation}
    \begin{aligned}
        p_\theta(\mathbf{x}^{t-1}|\mathbf{x}^t) = \mathcal{N}(\mathbf{x}^{t-1};\mu_\theta(\mathbf{x}^t,t),\sum\nolimits_\theta(\mathbf{x}^t,t))
        \label{equ:reverseProcess}.
    \end{aligned}
\end{equation}

The learning objective is obtained by thinking about the variational lower bound:
\begin{equation}
    \begin{aligned}
        \mathbb{E}  \left[-\log p_\theta(\mathbf{x}^0) \right] \leq \mathbb{E}_q \left[-\log \frac{p_\theta(\mathbf{x}^{0:T})}{q(\mathbf{x}^{1:T}|\mathbf{x}^0)}\right] 
         = \mathbb{E}_q \left[-\log p(\mathbf{x}^T) - \sum_{t\geq1}\log\frac{p_\theta(\mathbf{x}^{t-1}|\mathbf{x}^t)}{q(\mathbf{x}^t|\mathbf{x}^{t-1})} \right] = L
        \label{equ:variationalLowerBound}.
    \end{aligned}
\end{equation}

According to Ho \emph{et al.}~\cite{ho2020denoising}, this loss can be further decomposed as:
\begin{equation}
    \begin{aligned}
        \mathbb{E}_q \left[{\underbrace{D_{KL}(q(\mathbf{x}^T|\mathbf{x}^0) \| p(\mathbf{x}^T))}_{L_T}} \right.           \left. { + \sum_{k>1} \underbrace{D_{KL}(q(\mathbf{x}^{t-1}|\mathbf{x}^t,\mathbf{x}^0) \| p_\theta(\mathbf{x}^{t-1}|\mathbf{x}^t))}_{L_{t-1}}} \underbrace{-\log p_\theta(\mathbf{x}^0|\mathbf{x}^1)}_{L_0} \right]
        \label{equ:lossOne}.
    \end{aligned}
\end{equation}

The term $L_{t-1}$ plays a crucial role in training the network described in Eq. (\ref{equ:reverseProcess}) to perform a single reverse diffusion step. 
In addition, the availability of a closed-form expression for the objective is facilitated by the fact that $q(\mathbf{x}^{t-1}|\mathbf{x}^t,\mathbf{x}^0)$ is also a Gaussian distribution~\cite{ho2020denoising}.

The optimal approach for parameterizing the model involves predicting the cumulative noise $\epsilon_0$ that is added to the current intermediate data $\mathbf{x}^t$. Therefore, this can derive the following parameterization for the predicted mean $\mu_\theta(\mathbf{x}^t,t)$:
\begin{equation}
    \begin{aligned}
        \mu_\theta(\mathbf{x}^t,t) = \frac{1}{\sqrt{\alpha_t}}(\mathbf{x}^t-\frac{\beta_t}{\sqrt{1-\bar{\alpha}_t}}\epsilon_\theta(\mathbf{x}^t,t))
        \label{equ:predicteMean}.
    \end{aligned}
\end{equation}
Based on the $L_{t-1}$ term in Eq. (\ref{equ:lossOne}), Ho \emph{et al.}~\cite{ho2020denoising}  derive the following simplified training objective:
\begin{equation}
    \begin{aligned}
        L_\text{simple} = E_{t,\mathbf{x}^0,\epsilon} \left[\| \epsilon - \epsilon_\theta(\mathbf{x}^t,t)\|^2 \right]
        \label{equ:trainObjective}.
    \end{aligned}
\end{equation}
The method proposed in~\cite{dhariwal2021diffusion} can further improve the training and reasoning efficiency.


By leveraging the independent nature of the noise added at each step Eq. (\ref{equ:diffusionProcess}), the total noise variance can be calculated as $\bar{\alpha_t} = \prod^t_{s=1}(1-\beta_s)$. Thus,  the forward process can be represented as a single step:
\begin{equation}
    \begin{aligned}
        q(\mathbf{x}^t|\mathbf{x}^0) = \mathcal{N}(\mathbf{x}^t;\sqrt{\bar{\alpha}_t}\mathbf{x}^0,(1-\bar{\alpha_t})\mathbf{I})
        \label{equ:singleStep}.
    \end{aligned}
\end{equation}
Once trained, we can sample $\mathbf{x}_0$ from Eq. (\ref{equ:reverseProcess}) with a prior such as Gaussian noise.

\section{Related Work}




Initially, statistical methods are introduced to handle missing data by substituting missing values with statistics (e.g., zero values, mean values, last observed values) or using simple statistical models~\cite{amiri2016missing}. Furthermore, machine learning techniques, such as matrix factorization~\cite{chen2019bayesian, liu2012tensor, chen2020low} as well as Bayesian models and sparse network inference~\cite{fang2024bayotide, obata2024mining}, have gained prominence in the literature for addressing missing values in multivariate time series. Due to their remarkable capability to capture intricate dependencies in an end-to-end fashion, deep learning techniques have emerged as front-runners in time series imputation tasks. A growing array of deep neural networks is being employed to fill in missing values in time series datasets~\cite{han2019review}. The current landscape of deep learning-based imputation models can be primarily categorized into four groups: RNN-based, VAE-based, GAN-based, and Diffusion-based approaches.

\subsection{RNN-based Methods}
RNN, a powerful deep learning model for sequential data processing, performs exceptionally well in time series imputation tasks~\cite{fang2023dual}. It operates through a self-looping mechanism, incorporating time step information into the network architecture. This enables the network to consider information from preceding time steps while processing each time step. Hence, it is widely employed for time series data imputation. For example, Che \emph{et al.}~\cite{che2018recurrent} propose GRU-D, a gated recurrent unit (GRU) variant, which introduces a decay mechanism for the hidden states of GRU to handle missing data in time series classification problems. M-RNN~\cite{yoon2018estimating} fills in missing values using hidden states from a bidirectional RNN. Similar to the bidirectional structure of GRU-D, BRITS~\cite{cao2018brits} is designed to take into account the correlation among different channels for multivariate time series imputation. Inspired by the idea that missing parts can be imputed by sampling from the distribution of available data, deep latent variable approaches have also been explored for the imputation task~\cite{ma2018eddi}. Specifically, Rezende \emph{et al.}~\cite{rezende2014stochastic} estimate the conditional distribution of missing values based on the observed distribution. Sampling is then conducted via a Markov chain to perform data denoising and imputation. Mattei \emph{et al.}~\cite{mattei2018leveraging} extend this approach by improving the sampling strategy with Metropolis-within-Gibbs sampling. Li \emph{et al.}~\cite{li2023multi} present a multi-stage deep residual collaboration learning framework, Multi-DRCF, to tackle the complex task of repairing missing traffic data.
DGCRIN~\cite{kong2023dynamic} and GARNN~\cite{shen2023bidirectional} enhance the intermediate processes of bidirectional recurrent architectures using graph neural networks to improve imputation performance.

\subsection{VAE-based Models}
VAE is a generative model that combines the ideas of an autoencoder-decoder structure and a probabilistic graphical model, which allows for data generation and interpolation by learning the latent distributions of the data. VAE can learn the latent distributions of the time series data, which can be used to generate new time series samples and perform time series imputation~\cite{okafor2021missing}. 
To improve time series data imputation performance, various decomponents are designed and involved in the basic VAE model. 
For example, P-VAE~\cite{ma2018partial} incorporates a permutation invariant encoder and lower bound into VAE. MIWAE~\cite{mattei2019miwae} extends the importance-weighted autoencoder~\cite{burda2016importance} and maximizes a potentially tight lower bound of the log-likelihood of the observed data, and HI-VAE~\cite{nazabal2020handling} leverages an extension of the variational autoencoder lower bound. 
GP-VAE~\cite{casale2018gaussian} is a variation of the VAE architecture used for time series estimation with a Gaussian Process prior in the latent space. This prior is employed to facilitate embedding data into a smoother and more interpretable representation. 
TimeCIB~\cite{choi2024conditional} and ReCTSi~\cite{lai2024rectsi} introduce conditional information bottleneck and decoupled pattern learning, respectively, for data imputation. Moreover, graph neural networks have also been employed to capture the correlations between dimensions of the sequence for time series imputation. These works treat time series as graph sequences and reconstruct missing values using autoencoders~\cite{cini2021filling, marisca2022learning, wang2023networked}.

\subsection{GAN-based Approaches}
GAN, comprising a generator and a discriminator, is employed for time series interpolation. The generator generates missing or corrupted time series data to fill in missing values or correct noise, while the discriminator distinguishes real data from fake data generated by the generator. Through this process, GAN effectively generates realistic time series data~\cite{liu2019naomi,luo2018multivariate}.
Further, Luo \emph{et al.}~\cite{luo2018multivariate} proposes a GRU for Imputation (GRUI) to model the temporal information of incomplete time series. Both the generator and discriminator in their GAN model are based on GRUI. 
E2GAN~\cite{luo2019e2gan} presents an end-to-end method with two stages, which employs a GRUI-based autoencoder as its generator, reducing the difficulty of model training and improving estimation performance.
Liu \emph{et al.}~\cite{liu2019naomi} introduces a non-autoregressive model called NAOMI, consisting of a bidirectional encoder and a multi-resolution decoder. NAOMI is further enhanced through adversarial training. Richardson \emph{et al.}~\cite{richardson2020mcflow} develops a training strategy to simultaneously train a normalizing flow and a deterministic inference network for missing data completion.
Miao \emph{et al.}~\cite{miao2021generative} propose a conditional generator for the target time series based on predicted labels. To effectively capture the complex spatio-temporal patterns in traffic data imputation, SCL~\cite{qin2021network} and HSPGNN~\cite{liang2024higher} capture the essential spatio-temporal relationships when significant signal corruption occurs for precise imputation.
ST-LBAGAN~\cite{yang2021st} and C$^3$S$^2$-GAL~\cite{li2024self} explore bi-directional attention and cyclical constraints for the GAN to fill in missing
values of traffic data.


\subsection{Diffusion-based Techniques} 
Diffusion models~\cite{ho2020denoising}, emerging as powerful generative models with an impressive performance on various tasks, have been utilized for imputing multivariate time series. These methods initiate the imputation of missing values from randomly sampled Gaussian noise, which is then transformed into estimates of the missing values~\cite{tashiro2021csdi}. 
In this thread, Park \emph{et al.}~\cite{park2022neural} propose CSDE, which presents two new losses that can efficiently generate complex time series in the data space without additional networks.
CSDI~\cite{tashiro2021csdi} imputes the missing data through score-based diffusion models conditioned on observed data, exploiting temporal and feature correlations by a two-dimensional attention mechanism. 
TimeGrad~\cite{rasul2021autoregressive} conditions its autoregressive forecasting on the LSTM-encoded representation of the current time series to estimate future values regressively. CSDI~\cite{tashiro2021csdi} uses score-based diffusion models conditioned on observed data to impute missing values, leveraging temporal and feature correlations with a two-dimensional attention mechanism.
SSSD~\cite{alcaraz2023diffusionbased} applies the state space model~\cite{gu2022efficiently} as the denoising module of the diffusion models~\cite{kong2021diffwave} to conduct imputation. 
PriSTI~\cite{liu2023pristi} constructs a priori knowledge and interpolates spatio-temporal data based on extracted conditional features and geographic information. 
MIDM~\cite{wang2023observed} employs new noise sampling, addition, and denoising mechanisms, and employs several new techniques to ensure consistency between observations and missing values.
DA-TASWDM~\cite{xu2023density} incorporates dynamic temporal relationships into the denoising network for medical time series imputation.

\textbf{Difference.}
The aforementioned VAE-based and GAN-based methods often suffer from unstable training processes. For example, VAEs are prone to posterior collapse, while GANs frequently encounter mode collapse, both of which lead to inferior imputation performance. Notably, methods that incorporate graph neural networks to capture the correlations between time series channels often combine these with VAEs or GANs for data imputation, and thus, they face similar challenges. To address these issues, diffusion-based methods have been introduced to enhance training stability. However, they may cause  disharmonious boundaries between missing and known regions and result in information loss regarding long-term data. In contrast, we propose a weight-reducing injection strategy based on the diffusion model to alleviate boundary disharmony, and  introduce a multi-scale TemS4-based U-Net to effectively capture long-range dependencies.


\section{\modelname~Model}

\begin{figure*}
\centering{\includegraphics[width=100mm]
{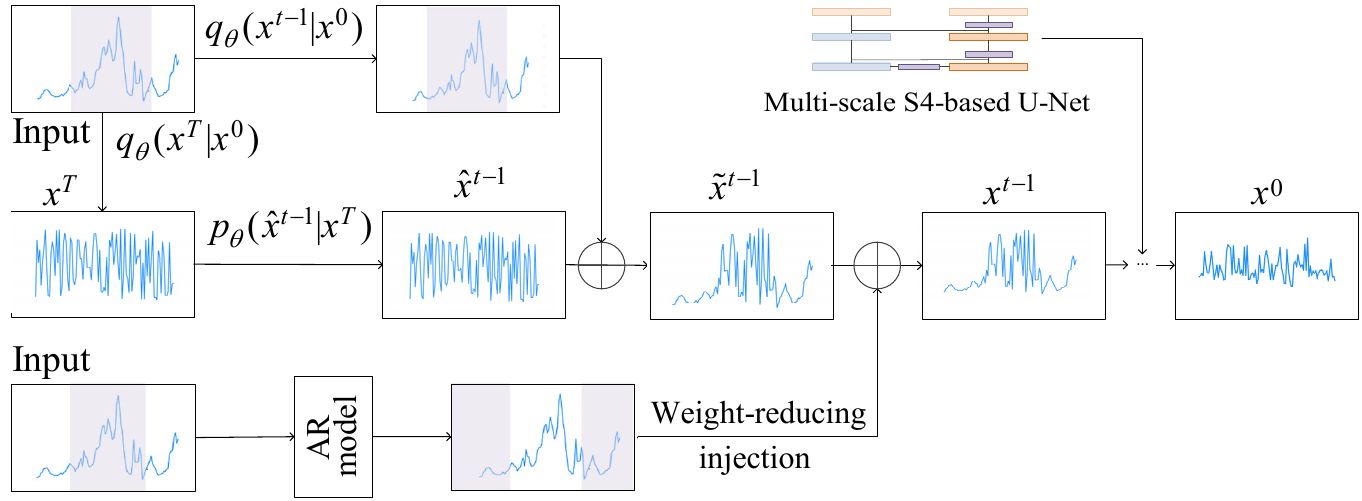}}
\caption{The \modelname~framework.}
\label{FigFrame}
\end{figure*}

The proposed \modelname~is composed of three components: the conditional diffusion imputation, the weight-reducing injection, and the multi-scale TemS4-based U-Net. An overview of these components is illustrated in Figure~\ref{FigFrame}. The conditional diffusion imputation estimates missing values by treating known values as conditions, which are exerted on the reverse diffusion process to generate estimated values. The weight-reducing injection strategy refines the generated values for missing points using the AR model, effectively mitigating boundary disharmony between missing and known regions. The multi-scale U-Net, built on temporal S4, leverages long-term dependencies to capture information beyond the missing regions, thereby enhancing data imputation.

\subsection{Conditional Diffusion Imputation using Known Values}

Due to the superior performance of diffusion models in data generation and imputation~\cite{lugmayr2022repaint,xiao2024diffusion,xiao2024counterfactual}, we employ DDPM as the basic imputation framework, utilizing known values as conditions to fill in missing values. Specifically, during the reverse generation process, we treat the known values as conditions exerted on the data generation process. Before conducting data imputation using reverse diffusion, we first train a denoising network, $\epsilon_\theta(\mathbf{x}^t,t,\mathbf{x}^0_\text{known})$, which is designed to predict the noise added in the $t$-th step. The training objective is as follows:
\begin{equation}
    \begin{aligned}
        L_\text{simple} = E_{t,\mathbf{x}^0,\mathbf{x}^0_\text{known},\epsilon} \left[\| \epsilon - \epsilon_\theta(\mathbf{x}^t,t,\mathbf{x}^0_\text{known})\|^2 \right]
        \label{equ:contrainObjective}.
    \end{aligned}
\end{equation}


Then, we feed the prior $\mathbf{x}^T$ deprived from the time series data containing missing values into the reverse diffusion module and generate a clean sample $\mathbf{x}^0$ through iterative denoising, i.e., $\mathbf{x}^T \rightarrow \cdots \mathbf{x}^t \rightarrow \mathbf{\hat{x}}^{t-1} \rightarrow \mathbf{\tilde{x}}^{t-1} \rightarrow \mathbf{x}^{t-1} \rightarrow \cdots \rightarrow \mathbf{x}^0$. 
Specifically, according to the Markov chain, the conditional reverse diffusion process predicts $\mathbf{\hat{x}}^{t-1}$ based on the generated latent variable in the previous step, $\mathbf{x}^t$, and the condition, $\mathbf{x}^0_{known}$. After adding the condition $\mathbf{x}^0_{known}$, the reverse process of the DDPM becomes: 
\begin{equation}
    \begin{aligned}
    p_\theta(\mathbf{\hat{x}}^{t-1}|\mathbf{x}^t,\mathbf{x}^0_\text{known})  = \mathcal{N}(\mathbf{\hat{x}}^{t-1};\mu_\theta(\mathbf{x}^t,t,\mathbf{x}^0_\text{known}), \tilde {\beta_t} I)
        \label{equ:conreverseProcess},
    \end{aligned}
\end{equation} 
where $\mathbf{x}^0_\text{known}$ denotes known values and $\Tilde{\beta_t}$ is a fixed constant. $\mu_\theta(\mathbf{x}^t,t,\mathbf{x}^0_{known})$ is the mean estimated based on the denoising network $\epsilon_\theta(\mathbf{x}^t,t,\mathbf{x}^0_\text{known})$:
\begin{equation}
    \begin{aligned}
        \mu_\theta(\mathbf{x}^t,t,\mathbf{x}^0_\text{known}) = \frac{1}{\sqrt{\alpha_t}}(\mathbf{x}^t-\frac{\beta_t}{\sqrt{1-\bar{\alpha}_t}}\epsilon_\theta(\mathbf{x}^t,t,\mathbf{x}^0_\text{known}))
        \label{equ:conpredicteMean}.
    \end{aligned}
\end{equation}
As a result, the reverse diffusion process can be expressed:
\begin{equation}
    \begin{aligned}
        \hat{x}^{t-1}=\frac{1}{\sqrt{\alpha_t}}(\mathbf{x}^t-\frac{\beta_t}{\sqrt{1-\bar{\alpha}_t}}\epsilon_\theta(\mathbf{x}^t,t,\mathbf{x}^0_\text{known})) + \sqrt{\Tilde{\beta_t}}\textbf{z}
        \label{equ:reknown}
    \end{aligned},
\end{equation}
where $\textbf{z}\sim(0, I)$ implying that each generation step is stochastic. This generative process iteratively refines the distribution until reaching a clean sample $\mathbf{x}^0$.
The forward process remains the same as that of the unconditional model, i.e., Eq.(\ref{equ:diffusionProcess}). 

\subsection{Weight-Reducing Injection on Missing Points}

The conditional diffusion model leverages known values to fill in missing data. However, during the early stages of reverse diffusion iterations, the generated values for the missing points may contain substantial noise, leading to disharmony at the boundaries between missing and known points. This disharmony can adversely affect the final imputation results~\cite{lugmayr2022repaint,shen2023non}. To address this issue, we propose a weight-reducing injection strategy to refine the generated values, improving their accuracy. This strategy first employs a linear AR model~\cite{rasul2021autoregressive} to predict an initial estimate for the missing points. The estimated values are then injected into the missing data generated by the diffusion model to mitigate disharmony. During this process, the strategy applies a gradually decreasing weight to the injected estimates, as the generated values become progressively closer to the true values with more diffusion iterations. The method consists of two key steps: missing value prediction and weight-reducing injection.

\textbf{Missing Value Prediction.} 
We introduce an AR model with the transformer structure to predict the missing values based on the known values as it can efficiently deal with time series data~\cite{rasul2021autoregressive,nie2023time}. Specifically, we first map the observed points to the latent space with dimension $d$:
\begin{align}
\mathbf{x}^{d}=\mathbf{W}_{p}{{x}^0_\text{known}}+\mathbf{W}_\text{pos},
\end{align}
where ${\mathbf{W}}_{p}$ denotes a trainable linear projection, $\mathbf{x}^0_\text{known}$ is the known value, and $\mathbf{W}_\text{pos}$ refers to a learnable additive position encoding that is applied to monitor the temporal order of observed points. 

Then, each head ${h}=1,...,H$ in multi-head attention will transform $\mathbf{x}^{d}$  into query matrices $Q_{h}$, key matrices $K_{h}$ and value matrices $V_{h}$, 
\begin{equation} \begin{aligned}
 Q_{h}&=({x}^{d})^{T}\mathbf{W}_{h}^{Q},\\
 K_{h}&=({x}^{d})^{T}\mathbf{W}_{h}^{K},\\
 V_{h}&=(\mathbf{x}^{d})^{T}\mathbf{W}_{h}^{V}.
\end{aligned} \end{equation}
After that, a scaled production is used for obtaining attention output $\mathbf{O}_{h}^{(i)}$:
\begin{equation}
(\mathbf{O}_{h})^T=\text{Attention}(Q_{h},K_{h},V_{h})=\text{Softmax}(\frac{Q_{h}K_{h}}{\sqrt{d_{k}}})V_{h}.
\end{equation}
The multi-head attention block also includes a BatchNorm layer and a feed-forward network with residual connections. Next, it generates the representation denoted as $\mathbf{z}$. Finally, a flattened layer with linear head is used to obtain the predicted result $\mathbf{z}_{ar}$. The entire process is presented as follows:
\begin{equation}
\begin{aligned}
\mathbf{O} &= \mathrm{Concat}(\mathbf{O}_1, \mathbf{O}_2, \ldots, \mathbf{O}_H) \mathbf{W}^O, \\
\mathbf{z} &= \text{FFN}(\text{BatchNorm}(\mathbf{O})), \\
\mathbf{z}_{ar} &= \text{Linear}(\text{Flatten}(\mathbf{z})),
\label{equ:armodel}
\end{aligned}
\end{equation}
where $\mathbf{W}^{O}$ is the linear transformation matrix used in the multi-head attention mechanism to concatenate and transform the outputs of the heads into the final output.

The AR model is pre-trained on the training set by minimizing the $\ell_2$-distance between $\mathbf{z}_{ar}$ and the ground-truth values: 
\begin{equation}
    \begin{aligned}
       \mathcal{L}_{ar} = \| \mathbf{x}^0_\text{known} - \mathbf{z}_{ar}\|^2
        \label{equ:larmodel}
    \end{aligned}.
\end{equation}
Note that although the AR model is an autoregressive model, all the columns of $\mathbf{z}_{ar}$ are obtained simultaneously rather than sequentially column by column. This avoids the problems of error accumulation and slow inference.

\textbf{Weight-Reducing Injection.} 
To address the disharmony between missing and known points, we inject the predicted values $\mathbf{z}_{ar}$, computed from Eq.\eqref{equ:armodel}, into the generated values $\Tilde{x}^{t-1}$, derived from Eq.\eqref{equ:reknown}, using a gradually reducing weight. In the early stages, the predicted values are assigned a larger weight during injection because the generated values from the reverse diffusion process deviate significantly from the true values, and the predicted values are more accurate at this stage. This injection with a larger weight helps to correct the deviation. Conversely, in the later stages, the predicted values are assigned a smaller weight, as the generated values of the missing points are expected to approach the true observed values more closely than the predicted ones. Consequently, applying a reduced weight in this phase encourages the generated values to converge toward the observed ones.

Specifically, we introduce a monotonic function $h(\cdot)$ to compute the injection weights. $h(\cdot)$ has a co-domain from 0 to 1 and can be obtained based on the exponential decay equation: 
\begin{equation}
    \begin{aligned}
        h(t-1)=1 - N_{0} e^{-\lambda(t-1)}
        \label{equ:weightfunction},
    \end{aligned}
\end{equation}
where $\lambda$ is a hyper-parameter representing the exponential decay constant, and $N_0$ is the initial quantity. As timestamp $t$ decreases and reverse sampling continues, the generated values of the missing points by the reverse diffusion are closer to the realistic values than the predicted values by the AR model. 
Accordingly, the disharmony between values of the missing points and known points gradually disappears, and the weight $h(\cdot)$ becomes smaller.

Having the weight function $h(\cdot)$, we mix $\mathbf{z}_{ar}$ computed by Eq.(\ref{equ:armodel}) and the generated values $\Tilde{x}^{t-1}$ computed by Eq.(\ref{equ:reknown}) to generate $\mathbf{x}^{t-1}$:
\begin{equation}
    \begin{aligned}
         \mathbf{x}^{t-1} = m \odot \mathbf{x}_\text{known}^{t-1} 
         + (1-m) \odot \bigg(h(t-1)\mathbf{z}_{ar}+\Big(1-h(t-1)\Big)\Tilde{x}^{t-1}\bigg), 
    \end{aligned}
 \end{equation}
where $m \odot \mathbf{x}_{known}^{t-1}$ and $(1-m) \odot (\cdot)$ denote the values of known points and the generated values of the missing points, respectively. 
$\mathbf{x}_\text{known}^{t-1}$ is noised known values computed  using the forward diffusion process:
    \begin{equation}
        \mathbf{x}_\text{known}^{t-1} \sim \mathcal{N}(\sqrt{\bar{\alpha}_t}\mathbf{x}^0, (1-\bar{\alpha}_t)\mathbf{I}). 
        \label{equKnownForward}
    \end{equation}
The weight-reducing method  exhibits several attractive properties:
(1) When $t\rightarrow T$, $h(t) \rightarrow 1$, thus the predictions by the AR model largely substitute the values generated by the reverse diffusion in the early steps.
(2) Due to the continuous property of the exponential function, it can correct deviation gently and progressively.

By introducing the AR model to modify and adjust the denoising process, the missing data can be estimated more accurately. 
This improves the quality of the generated data and generates the missing data which is closer to the ground-true values.

\begin{figure*}[!t]
    \centerline{\includegraphics[width=120mm]{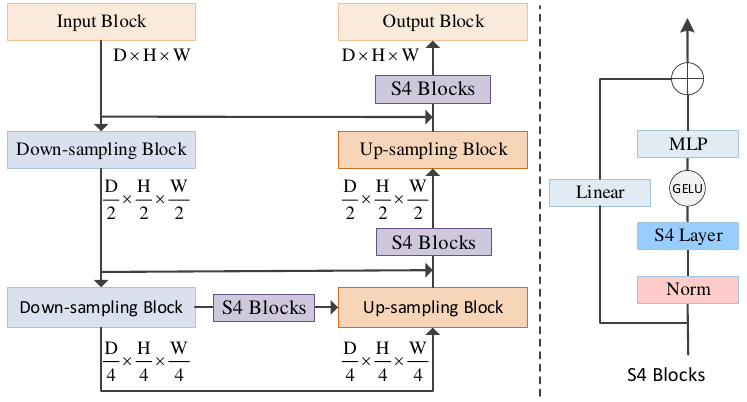}}
    \caption{The architecture of the  multi-scale TemS4-based U-Net.}
    \label{FigUnet}
\end{figure*}

\subsection{Multi-Scale TemS4-based U-Net}

In diffusion models, the U-Net~\cite{ronneberger2015u} based on a Wide ResNet~\cite{zagoruyko2016wide} is commonly employed as the denoising neural network~\cite{ho2020denoising}. However, components such as convolution and pooling operations in this U-Net backbone are not efficient at handling data with long-range dependencies~\cite{gu2022efficiently,gu2022parameterization}. This limitation becomes particularly significant when dealing with continuous missing points, as exploring long-term data information is crucial for capturing patterns beyond the missing point set~\cite{xiao2023imputation}. To address this, we design a multi-scale S4-based U-Net backbone to consolidate information from multiple tiers at different resolutions. This multi-scale S4-based U-Net is used as the denoising neural network $\epsilon_\theta(\cdot)$ in the diffusion model.


Figure~\ref{FigUnet} provides an overview of our multi-scale TemS4-based U-Net. The temporal S4 block is integrated into three key parts of the U-Net, where the data is processed by this block to enhance feature extraction. The temporal S4 block employs skip connections that link different layers, facilitating the flow of information across multiple scales and enabling the model to effectively leverage both local and long-range features. In addition, it captures hierarchical information across multiple resolutions, allowing it to identify patterns that occur over different time frames. This multi-scale approach strengthens the model's ability to recognize and relate long-term dependencies within the data.

Specifically, it consists of multiple blocks where each block operates on different resolutions and dimensions. Starting from a high resolution and dimension, the proposed model gradually decreases temporal resolution and channel dimension at each block. Because of this multi-scale strategy, different blocks can effectively learn features at different scales, which helps the model to learn complex temporal dependencies over long time series data.
Moreover, the temporal S4 block is defined as:
\begin{equation} \begin{aligned}
        y            & = \text{LayerNorm}(\mathbf{x}_{in}),   \\
        y            & = S4(y),                      \\
        y            & = \phi(y),                    \\
        y            & = \text{MLP}(y),              \\
        y_\text{out} & = y + \text{Linear}(\mathbf{x}_{in}),
        \label{equ:S4Block}
    \end{aligned} \end{equation}
where $\phi$ represents a non-linear activation function, such as Gaussian Error Linear Unit (GELU)~\cite{tsuchida2021avoiding}. The utilization of the multi-scale strategy enables different blocks to capture features at various scales, facilitating the learning of intricate temporal dependencies in multivariate time series data.

We present the pseudo-code for the imputation procedure of our designed model in Algorithm~\ref{alg:diffusion}. This algorithm computes the prior, predicted results, and noised known values using Eq.\eqref{equ:singleStep}, Eq.\eqref{equ:armodel}, and Eq.\eqref{equKnownForward}, respectively (Lines 1-3). The reverse diffusion process performs $T$ iterations (Lines 4-8), and during each iteration, the noise is removed from the data generated in the previous iteration (Line 6). Subsequently, conditions with varying weights are applied to the generated data for the next iteration (Line 7), with the weights adjusted dynamically across iterations. After completing all iterations, the clean data are returned (Line 9).


\begin{algorithm}[ht]
	\renewcommand{\algorithmicrequire}{\textbf{Input:}}
    \renewcommand{\algorithmicensure}{\textbf{Output:}} 	
    \caption{Imputation procedure of our weight-reducing diffusion model.}
	\label{alg:diffusion}
	\begin{algorithmic}[1]
		\REQUIRE Time series $\mathbf{x}^{0}_{known}$; Weight function $h(\cdot)$
        \ENSURE Generated values $\mathbf{x}^0$
        \STATE Obtain prior $\mathbf{x}^T$ based on $\mathbf{x}^{0}_{known}$ using Eq.\eqref{equ:singleStep}
        \STATE Calculate predicted result $\mathbf{z}_{ar}$ using Eq.\eqref{equ:armodel}
        \STATE Compute noised known values $\mathbf{x}_\text{known}^{t-1}$ using Eq.\eqref{equKnownForward}
        \FOR{ $t = T,..., 1$ }
                \STATE $z \sim \mathcal{N}(\textbf{0},\textbf{I})$ if t $>$ 1, else $\textbf{z} = \textbf{0}$
                
                \STATE $\tilde{x}^{t-1}=\frac{1}{\sqrt{\alpha_t}}(\mathbf{x}^t-\frac{\beta_t}{\sqrt{1-\bar{\alpha}_t}}\epsilon_\theta(\mathbf{x}^t,t,\mathbf{x}^0_\text{known})) + \sqrt{\Tilde{\beta_t}}\textbf{z}$

                

                \STATE $\mathbf{x}^{t-1} = m \odot \mathbf{x}_\text{known}^{t-1} + (1-m) \odot \bigg(h(t-1)\mathbf{z}_{ar}+\Big(1-h(t-1)\Big)\Tilde{x}^{t-1}\bigg)$
        \ENDFOR
        \STATE \textbf{return} $\mathbf{x}^0$
	\end{algorithmic}
\end{algorithm}

\section{Experimental Evaluation}

\subsection{Experimental Settings}


\noindent\textbf{Datasets}.
We evaluate the performance of our proposed DSDI on three widely used time-series datasets: the medical time series data (\textit{DACMI}~\cite{luo2022evaluating}), Electricity Transformer Data (\textit{ETT})~\cite{zhou2021informer}, and Beijing multi-site Air-Quality Data (\textit{AQI})~\cite{zhang2018mixup}. Table \ref{TabDatasets} provides detailed statistics of the three datasets.

\emph{DACMI} is derived from the MIMIC-III database of electronic health records~\cite{johnson2016mimic}, and is shared publicly to conduct the DACMI challenge~\cite{luo2022evaluating}. The DACMI data set contains 13 blood laboratory test values of 8267 patients admitted to intensive care units (ICUs).
The data set comes with missing value indices and corresponding ground truths. Each patient's lab work is recorded for a specific time point. The DACMI challenge organizer has excluded patients with fewer than ten-time points or follow-up visits. Some lab values have not been measured at the time of the visit, resulting in some natively missing data in the ground-truth data. We exclude those records with natively missing values because our experiments require ground truth for evaluation.

\emph{ETT} is collected using power transformers from July 1, 2016, to June 26, 2018. The time resolution is 15 minutes, resulting in 69,680 samples with no missing values.
Each sample consists of seven features, including oil temperature and six different types of external power load features. The data from the first 16 months (July 2016 to October 2017) are used as the training set. 
The subsequent 4 months (November 2017 to February 2018) are used as the validation set. The remaining 4  months (March 2018 to June 2018) are utilized for testing. 

\emph{AQI} incorporates hourly air pollution data collected from 12 monitoring locations in Beijing, 
covering a time range from March 1, 2013, to February 28, 2017, totaling 48 months of data. Each monitoring site measures 11 continuous time series variables, such as PM2.5, PM10, and sulfur dioxide. 
These variables are aggregated together, resulting in a total of 132 features in the dataset. The dataset contains 1.6\% missing values. 
We partition the dataset into training, validation, and testing sets. 
The training set consists of data from the first 28 months (March 2013 to June 2015). The validation set includes the subsequent 10 months (July 2015 to April 2016). The testing set is composed of the remaining 10 months (May 2016 to February 2017). 
To create time series data samples, we extract data from every 24-hour period as an instance. 

\begin{table}
  \small
  \centering
  \caption{The statistics of datasets.}
  \setlength{\tabcolsep}{1mm}
  \renewcommand\arraystretch{1.2} 
  \begin{tabular}{cccccc}
    \toprule
    Data &\ total samples &\ features &\ sequence length &\ missing rate &\ interval \\
    \midrule
    DACMI & 8266 & 13 & 48  & 7.72\% & irregular \\    
    ETT & 5,803 & 7 & 24 & 0\%  & 15 min \\
    AQI & 1,461 & 132 & 24 & 1.6\% & 60 min \\    
    \bottomrule
  \end{tabular}

  \vspace{-2mm}
    \label{TabDatasets}
\end{table}



\noindent\textbf{Baselines}.
To assess the effectiveness and investigate the superiority of our \modelname~model, we conduct a comparative analysis against several baseline methods, including RNN-based methods (e.g., {M-RNN}~\cite{yoon2018estimating} and {BRITS}~\cite{cao2018brits}), 
VAE-based approaches (e.g., {GP-VAE}~\cite{fortuin2020gp}, {D3VAE}~\cite{li2022generative}, PoGeVon~\cite{wang2023networked}, and CTA~\cite{wi2024continuous}), 
GAN-based methods (e.g., {GRUI-GAN}~\cite{luo2018multivariate}, {E2GAN}~\cite{luo2019e2gan}, {PC-GAIN}~\cite{wang2021pc} and {MDCGAN}~\cite{li2023efficient}), 
and diffusion model-based models (e.g., {CSDI}~\cite{tashiro2021csdi}, {SSSD}~\cite{alcaraz2023diffusionbased} and MTSCI~\cite{zhou2024mtsci}).
\begin{itemize}
        \item 
        {M-RNN}~\cite{yoon2018estimating}: 
        A multi-directional recurrent neural network that interpolates missing values as constant data flows and estimates them between data flows.
        \item
        {BRITS}~\cite{cao2018brits}: An efficient method for interpolating missing values in multivariate time series involves recursive dynamic modeling, which estimates missing values by leveraging feature correlations within the time series data.
        \item 
        {GP-VAE}~\cite{fortuin2020gp}: A sequential latent variable model used for dimensional reduction and data imputation. Nonlinear dimensionality reduction is accomplished using the VAE method with a novel structured variational approximation.
        \item
        {D3VAE}~\cite{li2022generative}: A bidirectional variational autoencoder with diffusion, denoising, and disentanglement capabilities. A coupled diffusion probability model is proposed to incorporate time-series data without introducing arbitrary uncertainty.
        \item
        {PoGeVon}~\cite{wang2023networked}: A position-aware graph enhanced imputation model that leverages a variational autoencoder to predict missing values over both node time series features and graph structures.
        \item 
        {CTA}~\cite{wi2024continuous}: A continuous-time autoencoder that encodes an input time  series sample into a continuous hidden path and decodes it to reconstruct and impute the input.
        \item
        {GRUI-GAN}~\cite{luo2018multivariate}: An improved gated recurrent unit with the GAN that simulates the temporal irregularities of incomplete time series.
        \item 
        {E2GAN}~\cite{luo2019e2gan}: By utilizing both the discriminative loss and mean squared error, it estimates incomplete time series by generating the closest complete time series at a certain stage.
        \item 
        {PC-GAIN}~\cite{wang2021pc}: An unsupervised missing data imputation method that leverages latent class information contained within the missing data to enhance imputation capability.
        \item 
        {MDCGAN}~\cite{li2023efficient}: A multi-discriminator conditional generative adversarial network that intelligently learns the characteristics of the distributed data sets to accurately impute missing values.
        \item
        {CSDI}~\cite{tashiro2021csdi}: A conditional score-based diffusion model that can leverage the correlations between observed values.
        \item 
        {SSSD}~\cite{alcaraz2023diffusionbased}: A tailored diffusion model that introduces structured state-space models for time series data imputation.  
        \item
        {MTSCI}~\cite{zhou2024mtsci}: A conditional diffusion model for multivariate time series consistent imputation that incorporate a complementary mask strategy and a mixup mechanism to realize intra-consistency and inter-consistency.        
\end{itemize}






\newcommand{\tabincell}[2]{\begin{tabular}{@{}#1@{}}#2\end{tabular}}
\begin{table*}[t]
\small
  \centering
  \caption{Performance comparison on the three datasets.}
  \label{TabResults}
  \resizebox{\textwidth}{!}{
    \begin{tabular}{@{}p{1.63cm}cccccccccc@{}}
\toprule
        \textbf{Method} 
        & \multicolumn{2}{c}{\textbf{DACMI(Point)}}    
        & \multicolumn{2}{c}{\textbf{DACMI(Block)}}
        & \multicolumn{2}{c}{\textbf{ETT(Point)}}    
        & \multicolumn{2}{c}{\textbf{ETT(Block)}}        
        & \multicolumn{2}{c}{\textbf{AQI(Simulated)}} 
        \\
\cmidrule(lr){2-3} \cmidrule(lr){4-5} \cmidrule(l){6-7} \cmidrule(l){8-9} \cmidrule(l){10-11} 
        & 
       \footnotesize{MAE} & \footnotesize{RMSE} &
       \footnotesize{MAE} & \footnotesize{RMSE} &
       \footnotesize{MAE} & \footnotesize{RMSE} & 
        \footnotesize{MAE} & \footnotesize{RMSE} &
       \footnotesize{MAE} & \footnotesize{RMSE}  \\ 
       
  \midrule
        M-RNN  &
        $0.481\pm0.008$ &
        $0.716\pm0.021$ & 
        $0.499\pm0.009$ &  
        $0.755\pm0.022$ & 
        $0.381\pm0.008$ & $0.434\pm0.029$ & $0.431\pm0.009$ & $0.625\pm0.030$ & $0.308\pm0.001$ & $0.671\pm0.019$ \\
        BRITS  &
        $0.213\pm0.003$ &
        $0.439\pm0.020$ &
        $0.234\pm0.003$ &
        $0.477\pm0.021$ & $0.134\pm0.002$ & $0.265\pm0.017$ & $0.198\pm0.005$ & $0.497\pm0.017$ & $0.167\pm0.001$ & $0.549\pm0.015$ \\
   \midrule
        GP-VAE &
        $0.325\pm0.004$ &
        $0.571\pm0.018$ & 
        $0.342\pm0.005$ &  
        $0.609\pm0.019$ & $0.281\pm0.004$ & $0.316\pm0.022$ & $0.436\pm0.006$ & $0.579\pm0.025$ & $0.285\pm0.003$ & $0.643\pm0.015$ \\
        D3VAE  &
        $0.203\pm0.004$ & 
        $0.453\pm0.018$ & 
        $0.228\pm0.004$ & 
        $0.522\pm0.019$ & $0.132\pm0.002$ & $0.258\pm0.017$ & $0.231\pm0.003$ & $0.477\pm0.019$  & $0.203\pm0.002$ & $0.584\pm0.010$ \\
        PoGeVon & 
        $0.187\pm0.003$ &
        $0.456\pm0.017$ &
        $0.197\pm0.003$ &
        $0.492\pm0.018$ & 
        $0.164\pm0.002$ &
        $0.275\pm0.015$ &
        $0.203\pm0.003$ &
        $0.435\pm0.018$ &
        $0.172\pm0.002$ &
        $0.502\pm0.013$ \\
        CTA & 
        $0.175\pm0.003$ &
        $0.438\pm0.015$ &
        $0.186\pm0.005$ &
        $0.471\pm0.017$ &
        $0.123\pm0.002$ &
        $0.249\pm0.014$ &
        $0.186\pm0.003$ & 
        $0.417\pm0.017$ &
        $0.153\pm0.002$ &
        $0.491\pm0.010$ \\
    \midrule
        GURI-GAN & 
        $0.745\pm0.006$ &
        $1.032\pm0.031$ &
        $0.764\pm0.008$ &
        $1.113\pm0.032$ & $0.619\pm0.005$ & $0.739\pm0.041$ & $0.823\pm0.007$ & $1.031\pm0.041$ & $0.805\pm0.003$ & $1.201\pm0.010$ \\
        E2GAN    &
        $0.665\pm0.005$ &
        $0.886\pm0.023$ &
        $0.682\pm0.007$ &
        $0.921\pm0.025$ & $0.591\pm0.004$ & $0.713\pm0.037$ & $0.797\pm0.006$ & $0.978\pm0.037$  & $0.766\pm0.001$ & $1.155\pm0.012$\\
        PC-GAIN  &
        $0.332\pm0.003$ &
        $0.577\pm0.017$ &
        $0.348\pm0.005$ &
        $0.617\pm0.019$ & $0.202\pm0.002$ & $0.318\pm0.019$ & $0.425\pm0.004$ & $0.598\pm0.021$ & $0.291\pm0.001$ & $0.677\pm0.015$ \\
        MDCGAN   &
        $0.186\pm0.002$ &
        $0.447\pm0.014$ &
        $0.199\pm0.003$ &
        $0.473\pm0.015$ & $0.121\pm0.002$ & $0.248\pm0.016$ & $0.207\pm0.003$ & $0.425\pm0.017$ & $0.155\pm0.001$ & $0.517\pm0.011$ \\
  \midrule
        CSDI    &
        $0.244\pm0.003$ &
        $0.478\pm0.017$ &
        $0.279\pm0.005$ &
        $0.516\pm0.018$ & $0.163\pm0.006$ & $0.292\pm0.022$ & $0.289\pm0.005$ & $0.497\pm0.020$ & $0.209\pm0.002$ & $0.559\pm0.015$ \\ 
        SSSD    &
        $0.202\pm0.003$ &
        $0.445\pm0.012$ &
        $0.215\pm0.003$ &
        $0.480\pm0.013$ & $0.119\pm0.002$ & $0.249\pm0.015$ & $0.234\pm0.002$ & $0.478\pm0.015$ & $0.157\pm0.001$ & $0.511\pm0.010$ \\
        MTSCI  &
        $0.171\pm0.003$ &
        $0.423\pm0.003$ &
        $0.183\pm0.003$ &
        $0.462\pm0.003$ &
        $0.117\pm0.002$ &
        $0.243\pm0.013$ &
        $0.184\pm0.002$ & 
        $0.411\pm0.013$ & 
        $0.146\pm0.001$ &
        $0.487\pm0.010$ \\
  \midrule
  \textbf{\modelname} &
  $\pmb{0.162\pm0.002}$ &
  $\pmb{0.411\pm0.010}$ &
  $\pmb{0.182\pm0.002}$ &
  $\pmb{0.445\pm0.012}$ & $\pmb{0.108\pm0.001}$ & $\pmb{0.237\pm0.012}$ & $\pmb{0.169\pm0.001}$ & $\pmb{0.399\pm0.012}$ & $\pmb{0.139\pm0.001}$ & $\pmb{0.473\pm0.009}$\\
  \bottomrule
\end{tabular}
}
\vspace{-3mm}
\end{table*}

\noindent\textbf{Experiment Setup}.
We use the Adam optimizer~\cite{kingma2014adam} with an initial learning rate of $3\times 10^{-6}$ and set the batch size to 16 for AQI and 32 for other datasets. We configure the diffusion model with 100 diffusion steps, balancing efficiency and effectiveness. The denoising module is trained for 120 epochs, and training is halted if the loss does not improve for a predefined number of epochs (e.g., 10). To evaluate performance, we utilize Mean Absolute Error (MAE) and Root Mean Square Error (RMSE) as evaluation metrics. MAE is the average of the absolute values of the errors, while RMSE is the square root of the mean of the squared differences between the predicted values and the actual observations. For DACMI and ETT, since the original datasets have a low number of missing values, we follow the approach in~\cite{cini2021filling,marisca2022learning} and consider two different missing patterns to artificially simulate missing values for evaluation: (1) Point missing, where 10\% of the available data is randomly dropped. (2) Block missing, where 5\% of the available data is randomly masked out, and a failure lasting $S \sim U(12, 48)$ time steps is simulated with a 0.15\% probability. For the AQI air quality dataset, we adopt the same evaluation strategy as previous work~\cite{yi2016st,liu2023pristi}, which simulates the distribution of real missing data.



\subsection{Performance Comparison}


Table \ref{TabResults} presents the imputation performance of various baseline methods, including point, block and simulated missing types. Lower values indicate better performance, and the best performance is highlighted in bold. We summarize the following observations: (1) Our \modelname~significantly outperforms all baseline methods across these datasets, particularly on the AQI dataset, where it achieves greater gains. This improvement is likely attributed to the higher number of features and the complex relationships within the AQI data, which increase the likelihood of boundary distortion during the imputation process. However, our proposed method effectively mitigates this distortion, leading to a substantial enhancement in imputation performance. (2) GAN-based methods (e.g., GRUI-GAN~\cite{luo2018multivariate} and E2GAN~\cite{luo2019e2gan}) perform much less effectively than other baseline methods. This could be due to the unstable training of GAN networks, where the loss values of the generator and discriminator cannot be effectively minimized. As a result, GAN-based methods may exhibit inferior performance compared to other baselines. (3) The diffusion model-based methods demonstrate more stable performance across all datasets. However, our \modelname~still surpasses these baselines, highlighting the effectiveness of our designed weight-reducing injection method and the multi-scale U-Net based on temporal S4. Furthermore, we evaluate \modelname's performance under hybrid missing scenarios, where the data has a 50\% probability of being masked using the point strategy and a 50\% probability of being masked using the block strategy. The results suggest that the performance under hybrid missing scenarios falls between that of the point missing and block missing cases. In addition, we calculate the runtime of our model on these three datasets, showing that its speed is competitive with typical diffusion-based models and is well-suited for large-scale datasets.


To compare imputation performance under different missing rates, we randomly mask data with varying missing rates ranging from 10\% to 90\%. We select the optimal method from each class of baseline methods (e.g., BEITS, D3VAE, MDCGAN, and SSSD) to compare against our \modelname~method. Figure \ref{DifferRate-MAE-RMSE} presents the comparison results. According to the figure, \modelname~outperforms these four baseline methods across all missing rates. This demonstrates that, with the support of the weight-reducing injection mechanism, \modelname~can better learn data features and achieve superior performance. Furthermore, while the performance of all methods declines as the missing rate increases, \modelname~exhibits a more gradual degradation trend compared to the other baseline methods. This suggests that the components designed for \modelname~effectively enhance its stability under varying missing rates.

\begin{figure}[!t]
    \centerline{\includegraphics[width=120mm]{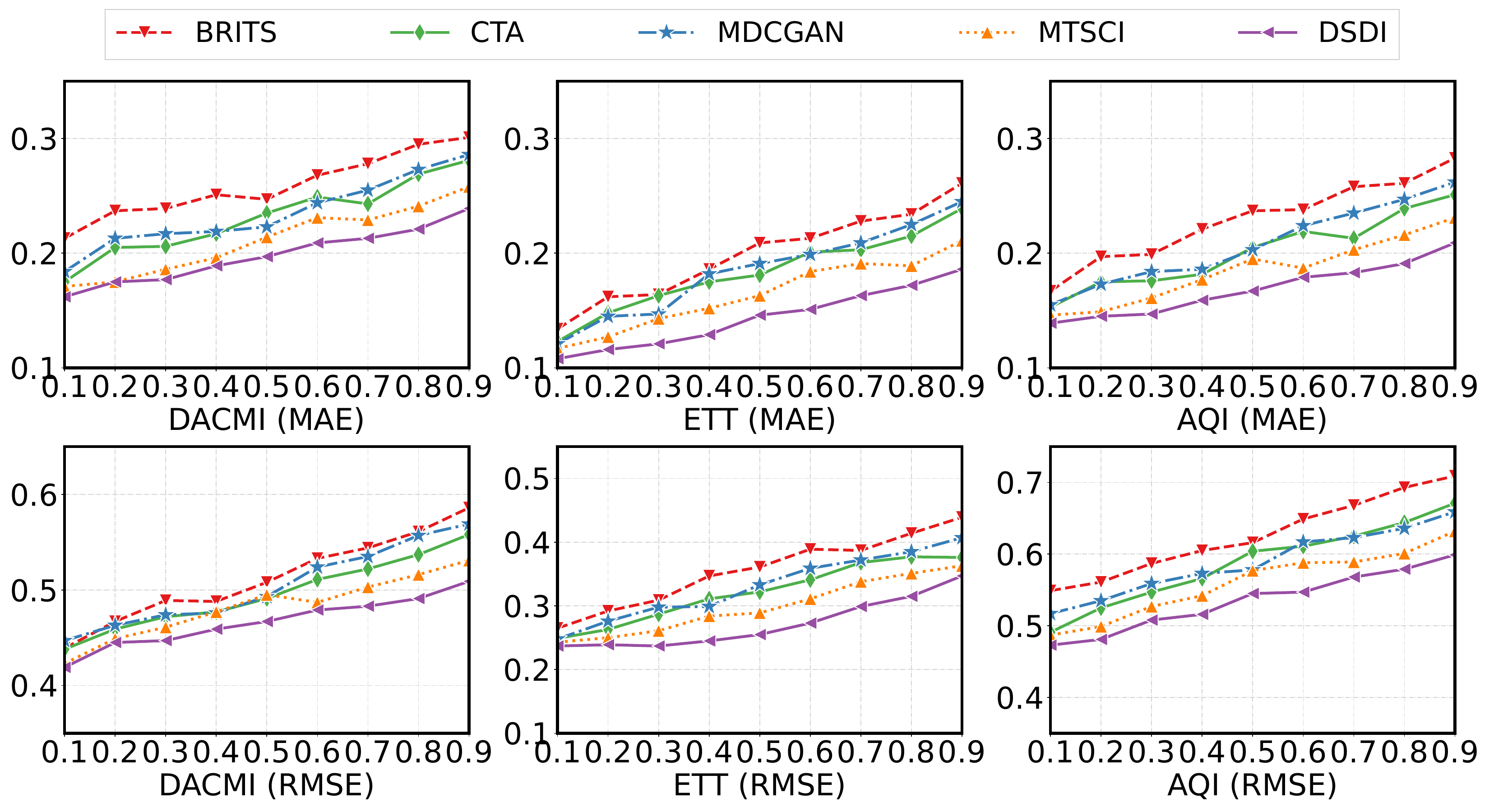}}
    \caption{MAE and RMSE at different missing rates.}
    \label{DifferRate-MAE-RMSE}
\end{figure}

{\begin{figure*}[!tb]    \centerline{\includegraphics[width=0.780\textwidth]{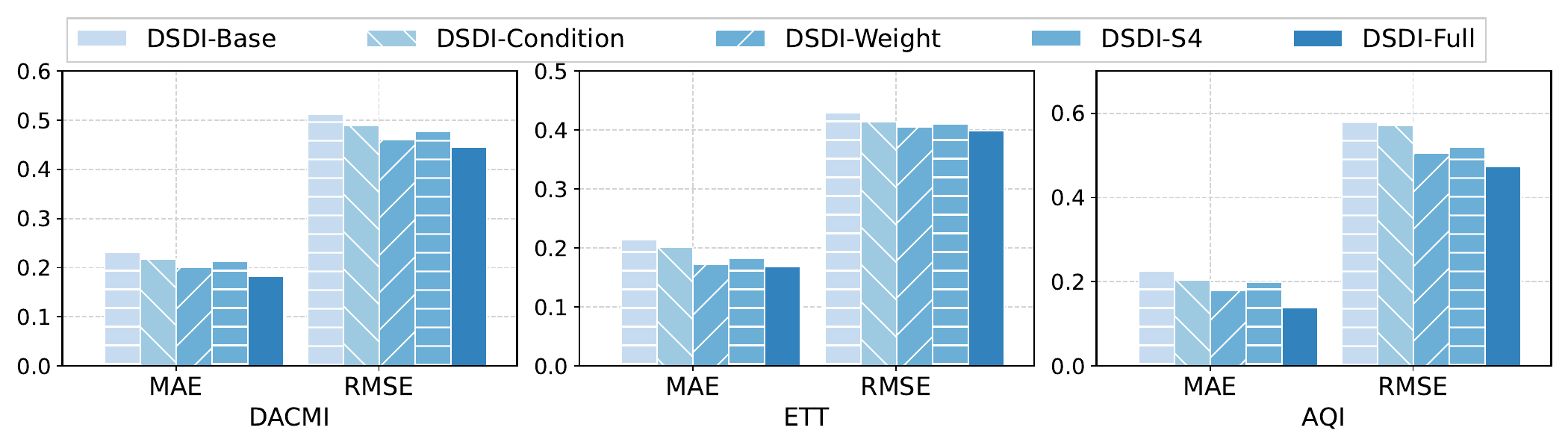}}
    \caption{The ablation study.}
    \label{FigAblationContinuous}
\end{figure*}}

\subsection{Ablation Study}

Here, we evaluate the role of the designed components in \modelname: the conditional diffusion mechanism, the weight-reducing injection strategy, and the multi-scale S4-based U-Net. We investigate the following variants: (1) \emph{\modelname-Base}: The basic diffusion model without any of the three components. (2) \emph{\modelname-Condition}: A model that incorporates known values as conditions applied during the reverse diffusion process. (3) \emph{\modelname-Weight}: A model that uses the weight-reducing injection strategy to inject predicted values for the unknown regions into the generated values with reducing weights. (4) \emph{\modelname-S4}: A model equipped with the multi-scale S4-based U-Net. (5) \emph{\modelname-Full}: A model that integrates all three components.

Figure \ref{FigAblationContinuous} presents the results of the variants on the three datasets. The figures indicate that the results of \modelname-Condition are superior to those of \modelname-Base, demonstrating that treating the known values as conditions effectively improves the estimation of missing values. Similarly, the results of \modelname-Weight outperform those of \modelname-Base, emphasizing the importance of the weight-reducing injection strategy. Among the three components, the performance improvement of \modelname-Weight is more significant than that of \modelname-Condition and \modelname-S4 in most scenarios. This highlights that the weight-reducing injection strategy is particularly effective for addressing distorted boundaries. This strategy ensures that boundaries transition gradually during the imputation process, which is crucial for achieving accurate estimations. Finally, the superior performance of \modelname-Full demonstrates the combined effectiveness of these components, resulting in a substantial improvement in the overall performance.

\subsection{Hyper-Parameter Selection}

Here, we examine the role of two crucial hyperparameters in \modelname. First, the number of iterations ($T$) in the diffusion model influences both the runtime and the quality of the imputed data. To evaluate its impact, we investigate different values of $T$, with the results presented in Figure \ref{MAE-HyperT}. As shown, the final imputation results improve as $T$ increases as first. However, when $T$ exceeds 100, the results stabilize and are comparable to those obtained at $T = 100$. Therefore, we chose $T = 100$ for the experiments to balance computational efficiency and imputation quality.

Second, in the weight-reducing injection strategy, the parameter $\lambda$ in Eq.(\ref{equ:weightfunction}) is used to adjust the speed of weight changes, which directly impacts the quality of the generated data. We evaluate the role of $\lambda$ in Figure~\ref{Fig:WeighHype}. This figure shows that $\lambda$ affects imputation performance to some extent. As the value of $\lambda$ increases, the performance deteriorates. This is likely because a larger $\lambda$ causes the weight to approach its maximum value of 1 more slowly. This conflicts with our design, where the predicted values should become less important in the later stages of the diffusion process.

\begin{figure}
\centering{\includegraphics[width=105mm]{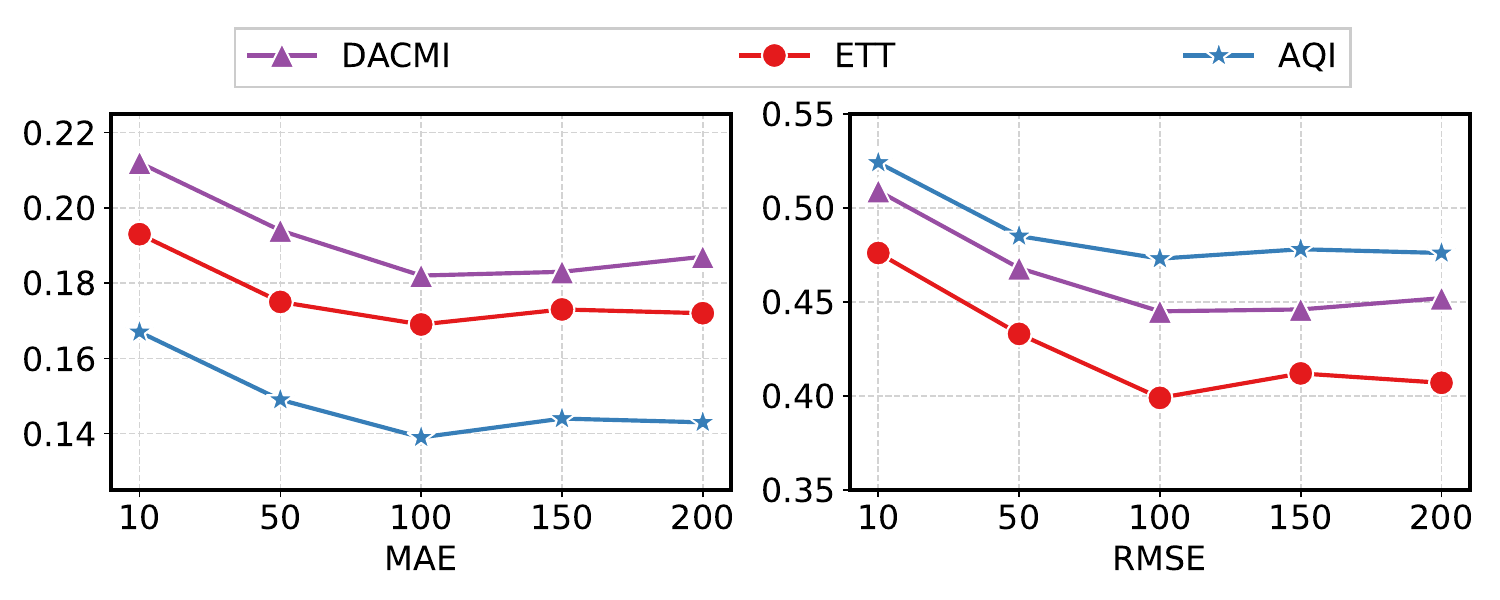}}
\vspace{-3mm}
\caption{The impact of the diffusion steps.}
\label{MAE-HyperT}
\vspace{-3mm}
\end{figure}

\begin{figure*}
\centering{\includegraphics[width=110mm]{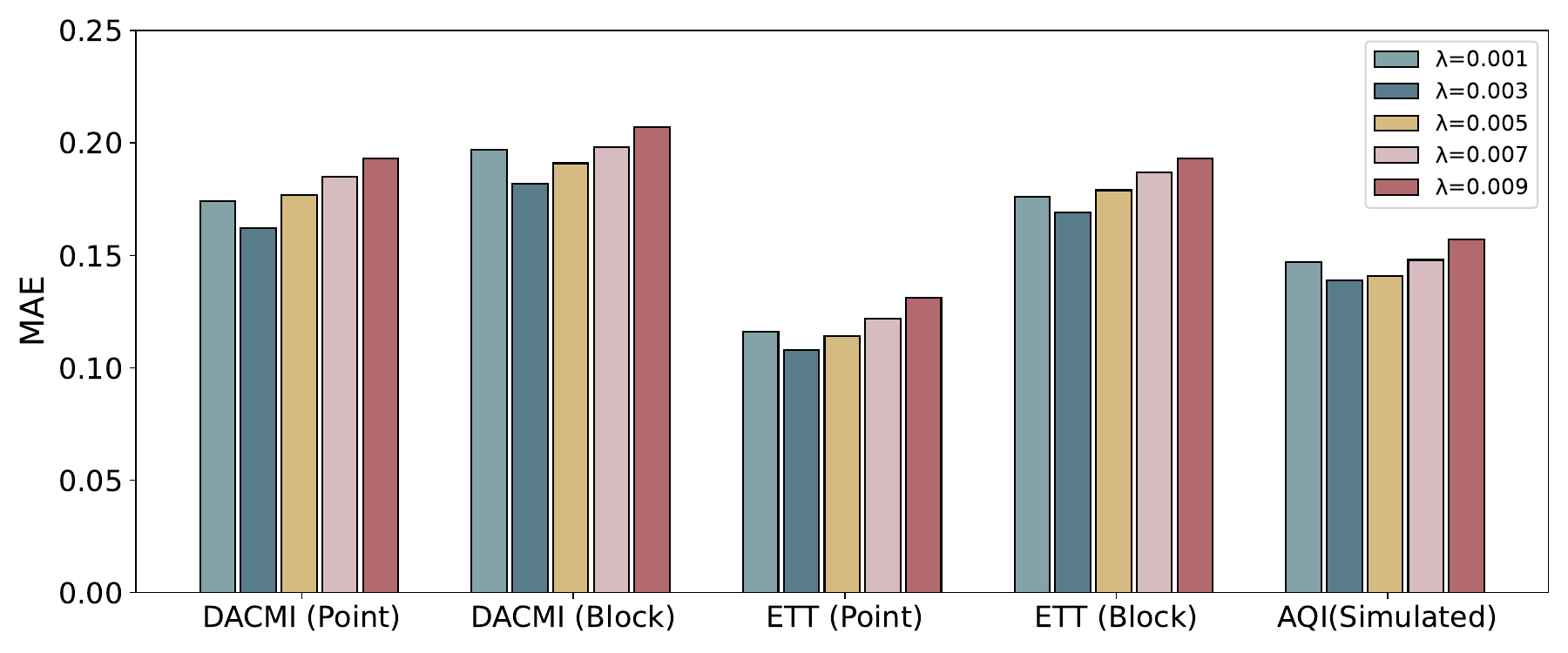}}
\caption{MAE for different weighting parameters $\lambda$ values.}
\label{Fig:WeighHype}
\vspace{-3mm}
\end{figure*}

\section{Conclusion}

In this work, we proposed a novel framework, DSDI, for time series imputation, which can effectively mitigate the performance degradation caused by the disharmony between missing and known regions. We introduced a weight-reducing injection strategy that applies progressively diminishing weights to the predicted values of missing points during the reverse diffusion process, gradually refining the estimated results. Furthermore, we developed a multi-scale TemS4-based U-Net to capture long-range dependencies, enhancing performance particularly in scenarios with consecutive missing data. Extensive experiments on real-world datasets demonstrate that DSDI outperforms state-of-the-art methods.


There are several promising directions for future work. One direction involves improving computational efficiency. While diffusion models produce highly realistic samples, their sampling process tends to be slower than that of other generative models. Enhancing computational efficiency could make diffusion models more practical for real-world applications. Another direction is to extend DSDI to various fields, such as finance and healthcare, where accurate data imputation is crucial for decision-making and predictive modeling. Finally, capturing uncertainty in estimating missing values represents a promising avenue for improving the robustness of the imputation model.







\bibliographystyle{elsarticle-num}
\bibliography{references}






\end{document}